\documentclass[letterpaper]{article} 
\usepackage{aaai2026}  
\usepackage{times}  
\usepackage{helvet}  
\usepackage{courier}  
\usepackage[hyphens]{url}  
\usepackage{graphicx} 
\usepackage{natbib}  
\usepackage{caption} 
\usepackage{algorithm}
\usepackage{amsmath}
\usepackage{amsfonts}
\usepackage{soul}
\usepackage{xcolor}
\usepackage{multirow}
\usepackage{booktabs}
\usepackage{siunitx}
\usepackage{subcaption}

\usepackage[noend]{algpseudocode}
\newtheorem{definition}{Definition}
\newtheorem{theorem}{Theorem}
\usepackage{newfloat}
\usepackage{listings}
\usepackage{pifont}
\DeclareCaptionStyle{ruled}{labelfont=normalfont,labelsep=colon,strut=off} 
\floatstyle{ruled}
\newfloat{listing}{tb}{lst}{}
\floatname{listing}{Listing}
\title{TiCAL:Typicality-Based Consistency-Aware Learning \\ for Multimodal Emotion Recognition}
\author{
    Wen Yin\textsuperscript{\rm 1}, Siyu Zhan\textsuperscript{\rm 1}, Cencen Liu\textsuperscript{\rm 1}, Xin Hu\textsuperscript{\rm 1},Guiduo Duan\textsuperscript{\rm 1,2}, Xiurui Xie\textsuperscript{\rm 1}, \\ Yuan-Fang Li\textsuperscript{\rm 3}, Tao He\textsuperscript{\rm 1,2}\thanks{Corresponding author.}\\
}
\affiliations{

    \textsuperscript{\rm 1}The Laboratory of Intelligent Collaborative Computing of UESTC \\
    \textsuperscript{\rm 2}Ubiquitous Intelligence and Trusted Services Key Laboratory of Sichuan Province \\
    \textsuperscript{\rm 3}Faculty of Information Technology, Monash University

    \{yinwen1999,zhansy,202411900402,202411900415,guiduo.duan,xiexiurui\}@uestc.edu.cn\\
    yuanfang.li@monash.edu, tao.he01@hotmail.com
}

\usepackage{bibentry}

\begin{document}
\maketitle
\begin{abstract}

Multimodal Emotion Recognition (MER) aims to accurately identify human emotional states by integrating heterogeneous modalities such as visual, auditory, and textual data. Existing approaches predominantly rely on unified emotion labels to supervise model training, often overlooking a critical challenge: \emph{inter-modal emotion conflicts}, wherein different modalities within the same sample may express divergent emotional tendencies. In this work, we address this overlooked issue by proposing a novel framework, Typicality-based Consistent-aware Multimodal Emotion Recognition (\textbf{TiCAL}), inspired by the stage-wise nature of human emotion perception. TiCAL dynamically assesses the consistency of each training sample by leveraging pseudo unimodal emotion labels alongside a typicality estimation. To further enhance emotion representation, we embed features in a hyperbolic space, enabling the capture of fine-grained distinctions among emotional categories. By incorporating consistency estimates into the learning process, our method improves model performance, particularly on samples exhibiting high modality inconsistency. Extensive experiments on benchmark datasets, e.g, MOSEI and MER2023, validate the effectiveness of TiCAL in mitigating inter-modal emotional conflicts and enhancing overall recognition accuracy, e.g., with about $2.6$\% improvements over the state-of-the-art DMD. 
\end{abstract}    
\section{1 Introduction}\label{sec:introduction}
Multimodal Emotion Recognition (MER) aims to infer human emotional states by integrating complementary information from multiple modalities such as visual expressions, acoustic signals, and textual content \cite{HumanMER2019,HumanMER2021,he2022towards,ucdver}. Compared to unimodal approaches \cite{SpeechEmotion,FacialEmotion},  which rely on a single source, MER benefits from cross-modal cues to achieve a strong understanding of affective behaviors.

\begin{figure}[t]
    \centering
    \includegraphics[width=0.95\columnwidth]{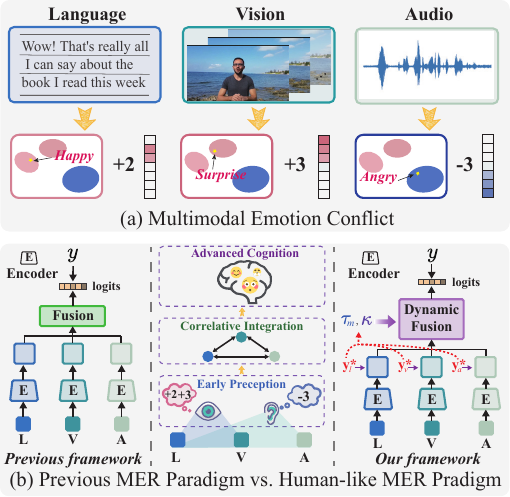}
    \caption{(a) Example of inter-modal emotional conflict where modalities show divergent affective cues.
(b) Comparison of our paradigm with prior work: we model modality-specific emotional tendencies and enforce consistency-aware dynamic fusion, inspired by human perception. $\tau_m$ and $\kappa$ represent typicality and consistency calculated using our method, respectively.
} 
    \label{fig:motivation}
\end{figure}

However, most existing MER methods assume that all modalities within a sample convey consistent emotional signals, supervising them with a single unified label  \cite{MSA_ECCV,DMD,CAGC}. This assumption is often invalid, as modalities can express conflicting emotional tendencies—a challenge we refer to as \emph{
inter-modal emotional conflict.} For example, Fig.~\ref{fig:motivation}(a) shows a MOSEI sample \cite{MOSEI} labeled as ``happy'', where the linguistic modality exhibits strong positive sentiment (polarity +$2$), while the acoustic modality reflects anger-like tendency (polarity -$3$). Such divergence arises from overlapping acoustic traits, e.g., high vocal intensity—shared by various emotions like happiness and anger. 

Treating such multimodal inputs as if they share a single emotion label (e.g., ``happy'') introduces ambiguity, leading to suboptimal feature learning and poor generalization. Although recent MER studies have improved fusion mechanisms through advanced encoders \cite{Emotionllama,VideoMAE,Videollama}  or contrastive objectives \cite{DMD,ConFEDE,CAGC}, they rarely address the fundamental cause of performance degradation—the absence of explicit modeling for inter-modal consistency and conflict. 
While \citet{emoconflict} attempts to address this by optimizing single-modal predictions, their approach does not capture inter-modal consistency, which is critical for effective fusion.


In contrast, human emotion perception is inherently stage-wise and consistency-aware. Neuroscience studies suggest that we first process unimodal cues independently (Early Perception \cite{2025EP1}), then integrate correlated signals (Correlative Integration     \cite{2018CI}), and finally reconcile conflicting information through high-level reasoning (Advanced Cognition \cite{2022AC}). Moreover, recent studies \cite{MMPareto,understandbias,Boostingcvpr2023} show that the degree of inter-modal consistency directly impacts the optimization of fusion networks—samples with conflicting signals require deeper and more adaptive integration. This raises a research question:
\emph{Can we explicitly quantify and leverage inter-modal consistency to achieve human-like, conflict-aware multimodal fusion for emotion recognition?}

To answer this, we propose \textbf{TiCAL} (\underline{T}yp\underline{i}cality-based \underline{C}onsistency-\underline{A}ware \underline{L}earning), a novel framework for multimodal emotion recognition (MER). TiCAL begins by generating intermediate pseudo-labels for each modality using high-confidence anchor samples list (HASL), which capture unimodal emotional tendencies. These pseudo-labels are then embedded into a hyperbolic space, providing a richer structure to model hierarchical and nuanced emotional cues. To ensure reliable cross-modal consistency, we introduce a typicality metric—inspired by the notion of typicality \cite{TAL}—to assess pseudo-label confidence and selectively emphasize trustworthy instances. Finally, we design a stage-wise perception framework that leverages both typicality and consistency, mimicking the human process of forming modality-specific impressions before integrating them into a robust multimodal understanding.

Experimental results on benchmark datasets, including CMU-MOSI, CMU-MOSEI, and MER2023, demonstrate that TiCAL provides an interpretable and robust solution for resolving inter-modal emotional conflicts in MER. 
Specifically, on CMU-MOSI, our method achieves state-of-the-art results with $\sim 2.6$\% improvement over strong baselines such as DMD \cite{DMD}.

In summary, our contributions are as follows:

\begin{itemize}
\item \textbf{Human-like Stage-wise Framework:} We propose \textbf{TiCAL} (Typicality-based Consistency-aware Learning), a novel framework that performs dynamic multi-stage fusion by leveraging inter-modal consistency and unimodal typicality,  mimicking human-like emotion perception.

\item \textbf{Reliable Consistency Estimation:} TiCAL captures unimodal emotional tendencies through pseudo-labels and introduces a typicality metric to assess their reliability, enabling robust and accurate consistency estimation across modalities.

\item  \textbf{State-of-the-art Performance:} We conduct extensive experiments on publicly available MER and MSA datasets, achieving superior performance across various SOTA models. Our results demonstrate the feasibility of emulating the human emotion perception mechanism and validate the effectiveness of multi-stage perception guided by consistent dynamics.

\end{itemize}

  

\section{2 Related Work}
\label{background}

\paragraph{Multimodal Emotion Recognition.}
Multimodal Emotion Recognition (MER) seeks to infer human emotional states by integrating signals from multiple modalities and has been widely used in image retrieval \cite{he2021semisupervised} and visual reasoning \cite{hu2025spade,yang2024towards}. Recent work has focused on developing advanced fusion strategies and interaction mechanisms to improve representation learning and model performance \cite{ConFEDE,CAGC,ou2025social,DMD,FDMER,PMR}. For instance, \citet{ConFEDE} introduced a unified contrastive learning framework that combines intra-sample modality decomposition with inter-sample supervised contrastive objectives. Other methods, such as \citet{CAGC} and \citet{DMD}, aim to disentangle modality-specific and modality-invariant features to facilitate more robust multimodal representations.
Although a recent study \cite{emoconflict} optimized multimodal heterogeneous information by minimizing unimodal losses, it still failed to effectively quantify and leverage inter-modal consistency, leading to suboptimal modality fusion and impaired branch optimization. 

\paragraph{Imbalanced Multimodal Learning.}
Imbalanced Multimodal Learning addresses the challenge of integrating diverse modalities that offer complementary yet asymmetrically signals. A common issue in this setting is the model’s tendency to overfit or over-rely on dominant modalities. To mitigate this, prior work has proposed various strategies to balance the learning process across modalities \cite{understandbias,dai2025unbiased,zhou2024lidarptq,PMRebalance,Boostingcvpr2023,li2024instant3d,he2021semantic,li2025uni}. For example, \citet{hu2024mvctrack} and \citet{Boostingcvpr2023} identified gradient conflicts in multitask-like training and proposed adaptive gradient modulation techniques to resolve them. \citet{MLA} approached the problem by introducing alternating single-modal optimization, enabling independent updates for each modality.
While effective, these methods primarily operate at the gradient level, focusing on adjusting optimization dynamics without explicitly modeling unimodal learning objectives. Even recent approaches such as \citet{PMRebalance} and \citet{Reconboosticml2024}, which refine objectives via prototype-based or regularization-based techniques, largely overlook the significance of the inter-modal consistency in guiding the different modal fusion stages.

\section{3 Methodology}
\begin{figure*}[ht]
    \centering
    \includegraphics[width=0.95\textwidth]{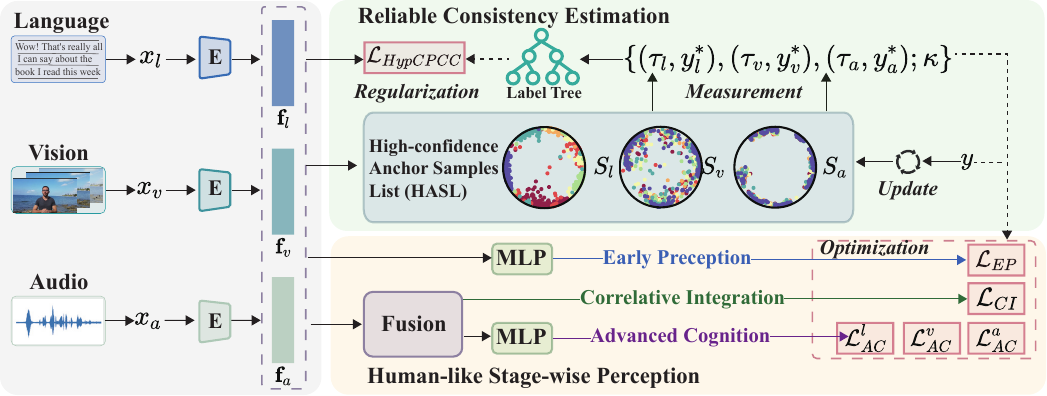}
    \caption{The overview of our proposed TiCAL. TiCAL comprises  Reliable Consistency Estimation and Human-like Stage-wise Perception.
    Specifically, we generate unimodal pseudo labels $y_m^*$ using a high-confidence anchor samples list (HASL) and regularize them in the hyperbolic space (In \S~3.1). Then we estimate the reliable inter-modal consistency $\kappa$ based on typicality $\tau_m$ and unimodal pseudo labels $y_m^*$ (In \S~3.2).
    Inspired by human emotion perception, we design a three-stage prediction structure—\textit{EP},\textit{CI}, and \textit{AC}—and conduct dynamic typicality-based consistency-aware optimization (In \S~3.3).}  
    \label{fig:framework}
\end{figure*}
\label{Methodology}

This section presents our proposed framework, \textbf{TiCAL}, as illustrated in Fig. \ref{fig:framework}.  In Sec.~3.1, we generate pseudo-labels for each unimodal input and regularize the corresponding emotion features in a hyperbolic space. This encourages the learning of modality-specific emotional cues while preserving the inherent hierarchical structure of emotional representations. In Sec.~3.2, we introduce a consistency metric to quantify inter-modal consistency for each sample, leveraging the pseudo-labels and a notion of typicality.  Finally, in Sec.~3.3, we draw inspiration from human emotional perception to adaptively fuse multimodal features. 


\paragraph{Problem Formulation.}
In the MER task, we consider a triplet of input modalities $\mathbf{x} = [x_m]$, where $m \in {l, v, a}$ denotes the language, visual, and acoustic modalities, respectively. Each input $x_m$ is processed by a modality-specific encoder to extract a corresponding feature representation $\mathbf{f}_m$. The resulting unimodal features are then integrated through a multimodal fusion module, which outputs the final emotion prediction $y$.


\subsection{3.1 Unimodal Labeling and Feature Structuring}\label{sec:UnimodalEmotion}
We consider generating the pseudo unimodal label for modality-specific emotion by anchor samples that the model predicts correctly with high confidence. Subsequently, we incorporate a hierarchically structured label into hyperbolic space to for unimodal features regularization. 

\subsubsection{Generating Pseudo Unimodal Labels from HASL}


A naïve approach to generating pseudo unimodal labels is to mask two modalities and predict from the remaining one. However, this often yields noisy labels due to the lack of multimodal context. Inspired by \citet{TAL}, we instead maintain a  high-confidence anchor samples list (HASL) to estimate pseudo labels via feature similarity in hyperbolic space. We define an HASL for each modality as
\begin{equation}
S_m = \left\{ (\mathbf{f}_{m_i}, y) \right\},
\end{equation}
where $\mathbf{f}_{m_i}$ denotes the feature of the $i$-th sample in modality $m$. The length $n$ of HASL is fixed and can be adjusted as a hyperparameter. During early training ($\text{epoch} \leq \lambda$), we populate $S_m$ with correctly predicted samples ($\hat{y}_i = y$) having confidence $u_{\hat{y}_i} > \theta$. After initialization, $S_m$ is updated using a First-In-First-Out (FIFO) strategy.
That is, the high-confidence samples saved earlier during training will be the first to be removed.

After $\lambda$ epochs, we assign pseudo unimodal labels by measuring feature similarity. Specifically, for a given sample with multimodal feature $\mathbf{f}_m$, we compute its distance to all stored features $\{\mathbf{f}_{m_j}\}^{S_m}_{j =1}$ in the HASL. The label of the closest anchor is treated as the pseudo label:
\begin{equation}\label{eq:d_min_pseudo_label}
    d_m= \min_{\substack{\mathbf{f}_{m_j} \in S_m}}d\left( \mathbf{f}_{m}, \mathbf{f}_{m_j}\right); y^*_m = y_{min},
\end{equation}
where $y_{min}$ represents the label with the minimum distance from the sample feature in HASL, $d(\cdot, \cdot)$ is any distance metric in the feature space.  We interpret $y^*_m$ as the unimodal emotional tendency of the sample inferred via similarity in representation space.


\subsubsection{Hyperbolic Regularization for Hierarchical Emotion Features}\label{sec:HypRegularization}
As discussed, emotional states often share overlapping features—e.g., both happiness and anger may involve heightened vocal intensity—leading to misjudgments of pseudo-labels. Thus, to generate high-quality pseudo labels $y_m^*$, we enhance their emotional discriminability through fine-grained unimodal representation learning \cite{simemotion1,simemotion2}. Specifically, we embed features in a hyperbolic space using the Poincaré Ball model \cite{ball}, which effectively captures the hierarchical and fine-grained structure of emotional states \cite{hyperbolic}.


\begin{theorem}[Poincaré Ball Model]
The Poincaré Ball Model \cite{ball} is a classical formulation of hyperbolic space in non-Euclidean geometry. It represents a $d$-dimensional hyperbolic space as an open unit ball: $\mathbb{B}^d = \left\{ \mathbf{x} \in \mathbb{R}^d \mid |\mathbf{x}| < 1 \right\}$, where $|\cdot|$ denotes the Euclidean norm. Given two points (e.g., feature vectors) $\mathbf{f}_1, \mathbf{f}2 \in \mathbb{B}^d$, their hyperbolic distance is defined as:
\begin{equation}\label{eq:d_hyperbolic}
d_{\mathbb{B}}(\mathbf{f}_1, \mathbf{f}_2) = \operatorname{arcosh} \left( 1 + 2 \frac{|\mathbf{f}_1 - \mathbf{f}_2|^2}{(1 - |\mathbf{f}_1|^2)(1 - |\mathbf{f}_2|^2)} \right).
\end{equation}
\end{theorem}

For pseudo label generation from HASL, we adopt the hyperbolic distance (Eq.~\ref{eq:d_hyperbolic}) to compare features and further optimize unimodal representations in hyperbolic space. To incorporate emotion hierarchies, we construct a weighted tree structure over emotion classes, enhancing one-hot labels into hierarchical ones. This structure captures class relationships more effectively. For details on the tree construction, see Appendix 1.1. During training, we apply a hyperbolic regularization that aligns feature embeddings with their corresponding hierarchical labels by minimizing the distance between feature-label node pairs in the tree.



\begin{definition} [Tree Node Pair Distance]
Given a weighted tree $\mathbb{T} = (\mathcal{V}, \mathcal{E}, \mathcal{W})$, where $\mathcal{V}$ is the set of nodes, $\mathcal{E}$ the set of edges, and $\mathcal{W}$ the corresponding edge weights, the tree distance $d_{\mathbb{T}}(v, v')$ is defined as the total weight of the shortest path connecting two nodes $v, v' \in \mathcal{V}$ in $\mathbb{T}$.
\end{definition}


To ensure that the unimodal features is empowered with the hierarchical structure, we compute tree-based distances $d_{\mathbb{T}}$ between label pairs in a batch. Simultaneously, we measure the hyperbolic distances $d_{\mathbb{B}}$ between corresponding feature representations in the Poincaré Ball model. To assess the alignment between the label hierarchy and the features, we employ the Hyperbolic Cophenetic Correlation Coefficient (HypCPCC) \cite{sokal1962comparison}, which quantifies the correlation between tree-based and hyperbolic distances as:
\begin{equation}\label{eq:cpcc}
    \text{HypCPCC} =\frac{
\sum_{i<j} \left( d_{\mathbb{T}} - \bar{d}_{\mathbb{T}} \right) \left( d_{\mathbb{B}} - \bar{d_{\mathbb{B}}} \right)
}{
\sqrt{ \sum_{i<j} \left( d_{\mathbb{T}} - \bar{d}_{\mathbb{T}} \right)^2 
\cdot \sum_{i<j} \left( d_{\mathbb{B}} - \bar{d_{\mathbb{B}}} \right)^2 }
},
\end{equation}
where $\bar{d}_{\mathbb{T}}$ and $\bar{d}_{\mathbb{B}}$ denote the mean distances across all node pairs in the batch for the tree and hyperbolic spaces, respectively. By maximizing the HypCPCC value during training, we encourage the unimodal features to respect the hierarchical relationships encoded by the pseudo labels, thus enhancing their fine-grained emotional discrimination capability.


\subsection{3.2 Reliable Inter-modal Consistency Estimation}\label{sec:ConsistencyMeasure}
In this section, we quantify the reliable inter-modal consistency of each sample based on pseudo unimodal labels and their associated typicality. To assess the reliability of each pseudo label, we compute a typicality score that reflects its alignment with the anchor distribution. We hypothesize that inter-modal consistency captures the degree of discrepancy across modalities, which directly influences prediction quality. Moreover, we conduct an analytical study  to validate this hypothesis  in Appendix 3.3 \& 3.4.

\subsubsection{Assessing Typicality for Pseudo Unimodal Labels}

Although pseudo unimodal labels can be derived via Eq.~\ref{eq:d_min_pseudo_label}, their reliability may vary. To assess this, we adopt typicality  \cite{beyondconfidence,TAL}, which suggests that typical samples are semantically representative, easy to learn, and generalize well. These samples tend to resemble the majority in structure and behavior.
We define typicality based on the relative hyperbolic distance to anchor samples  within a batch. Intuitively, samples closer to anchors are more typical and thus more trustworthy.

\begin{definition}[Unimodal typicality] 
We define the typicality of unimodal feature $\mathbf{f}_m$ as \footnote{Here typicality differs from ``typical distribution samples'' in probability theory \cite{beyondconfidence}, which refers to `` more valuable samples'', but we focus on the reliability of the estimation labels of different samples in batches.}
\begin{equation}
\label{eq:typicality}
    \tau_m = \frac{\max D_m - d_m}{\max D_m - \min D_m}, \footnote{Different from \citet{TAL}, we directly calculate hyperbolic distances based on features $\mathbf{f}_m$ instead of the mean $\mu$ and variance $\sigma^2$. More analysis can be found in the Appendix 3.1.3.}
\end{equation}
where $d_m$ is the sample’s minimum distance to HASL, and $D_m$ is the set of all such distances in the current batch. Here, $\max$ and $\min$ refer to the maximum and minimum values in $D_m$, respectively.
\end{definition}

A high typicality score $\tau_m$ indicates that the pseudo unimodal label is reliable, while a low $\tau_m$ suggests the sample is atypical or ambiguous. At this stage, we obtain a set of pseudo-label and typicality pairs for each modality: $\{ y_m^*, \tau_m \}$.

\subsubsection{Typicality-based Consistency Quantification}\label{sec:ConsistencyMetric}

In this work, we define inter-modal consistency by jointly considering pseudo labels and typicality, as relying on pseudo labels alone may overlook critical discrepancies. For instance, even if pseudo labels across modalities are aligned, large variations in typicality—where some modalities are highly typical while others are ambiguous—suggest underlying inconsistency in emotional expression. Specifically,  we define the calculation of inter-modal consistency  as: 

\begin{definition}[Inter-modal Consistency]
    With the pseudo unimodal labels and its corresponding typicality ($y_m^* ,\tau_m $), we define the inter-modal consistency as
\begin{equation}\label{eq:consistency}
    \kappa = \sqrt{(\tau_l \tau_v \tau_a )^t \cdot e ^{-k\cdot d_{label}}},
\end{equation}
where $t$ and $k$ are temperature and scaling hyperparameters. And $d_{label}$ is to measure the discrepancy between the estimated labels, defined as
\begin{equation}\label{eq:d_label}
    d_{label} = \left( \frac{|y_l^*-\mu_{label}|+|y_v^*-\mu_{label}|+|y_v^*-\mu_{label}|}{3} \right)^\rho,
\end{equation}
where $\mu_{label}$ is the mean value of estimated labels. $\rho$ is a density hyperparameter.
\end{definition}

\subsection{3.3 Human-Like Stage-wise Perception for MER}\label{sec:Optimization}
Inspired by human-like emotion processing, we decompose our model into three stages—Early Perception (EP), Correlative Integration (CI), and Advanced Cognition (AC)—guided by neuroscience insights (see Fig.~\ref{fig:framework}). EP performs unimodal emotion prediction from raw features; CI fuses multimodal information; and AC conducts more in-depth predictions. Each stage is supervised with the class-weighted cross-entropy losses $\mathcal{L}$ to alleviate the long-tailed problem in datasets. 
We design lightweight network structures, such as MLP, to ensure high portability. Implementation details are provided in the Appendix 2.
Specifically, we integrate the typicality (Eq.~\ref{eq:typicality}) and consistency (Eq.~\ref{eq:consistency}) metrics into the training objective, dynamically modulating optimization across stages. 
\paragraph{Unbaised Unimodal Weights with $\boldsymbol{\tau}$.}
Based on the typicality in Eq.(\ref{eq:typicality}), we can determine which modality can be dominant. To mitigate the risk of over-relying on any single dominant modality under conflicts or unreliable channels, we introduce  unbiased unimodal weights that assign lower relative weights to more typical (and hence dominant) modalities  as:
\begin{align}\label{eq:acloss}
    \mathcal{L}_\mathrm{AC} \!=&\! \varphi(\tau_l) \mathcal{L}_{AC}^l(\hat{y}, y) + \varphi(\tau_v)\mathcal{L}_{AC}^v(\hat{y}, y) \nonumber+ \\&\varphi(\tau_a) \mathcal{L}_{AC}^a(\hat{y}, y),
\end{align}
where $\varphi(\tau) = e^{(1 - \tau)}$ is a weight adjustment function that inversely scales with the typicality score.
\paragraph{Dynamic Optimization  with $\boldsymbol{\kappa}$.}
Prior work \cite{MMPareto,understandbias,Boostingcvpr2023} has shown that samples with high consistency should be prioritized during early training, while more inconsistent samples benefit from later-stage refinement. Following this principle, we dynamically weight the losses for the EP and AC stages based on the inter-modal consistency score $\boldsymbol{\kappa}$. The CI stage is assigned a fixed weight in order to ensure the basic integration of the modal. The overall task loss is defined as:
\begin{align}
\label{eq:taskloss}
\mathcal{L}_{\text{task}} = \kappa \mathcal{L}_{\text{EP}}(\hat{y}, y) + \mathcal{L}_{\text{CI}}(\hat{y}, y)  + (1 - \kappa) \mathcal{L}_{\text{AC}}(\hat{y}, y),
\end{align}
where $\hat{y}$ is the predicted label and $y$ is the ground truth. The EP stage is trained directly on original features, while the CI stage uses a cross-attention for basic integration and prediction (see Fig. \ref{fig:framework}). High-consistency samples ($\kappa\rightarrow1$) emphasize learning in the EP stage, while low-consistency samples ($\kappa\rightarrow0$) shift focus toward the AC stage.

\paragraph{The Overall Optimization.} 

In the early stage of training ($epoch \leq \lambda$), we only train the model with the task loss $\mathcal{L}_{task}$ without dynamic optimization by $\kappa$ and $\tau$, but in the later stage of training ($epoch > \lambda$), we introduce hyperbolic regularization to optimize the original unimodal features as:
\begin{equation}\label{eq:allloss}
    \mathcal{L}_{all} = \mathcal{L}_{task} - \mathcal{L}_\mathrm{HypCPCC}(\mathbf{f}_m, y_m^*).
\end{equation}

\paragraph{The Model Inference.} 
During the model inference stage, we can obtain three stages of emotion predictions ($\hat{y}_{EP}$, $\hat{y}_{CI}$, and $\hat{y}_{AC}$) and multimodal consistency $\kappa$. The final prediction label of the model is
\begin{equation}\label{eq:finalpred}
\hat{y}_{\text{final}} = \kappa \hat{y}_{\text{EP}}+ \hat{y}_{\text{CI}} + (1 - \kappa) \hat{y}_{\text{AC}}.
\end{equation}
\section{4 Experiments}
\label{sec:Experiments}

\subsection{4.1 Experimental Settings}
\paragraph{Datasets.}
We evaluate TiCAL on four widely used multimodal datasets: CMU-MOSI \cite{MOSI} and CMU-MOSEI \cite{MOSEI} for the Multimodal Sentiment Analysis (MSA) task, and DFEW \cite{DFEW} and MER2023 \cite{Mer2023} for the Multimodal Emotion Recognition (MER) task. \textbf{CMU-MOSI} comprises $2,199$ short monologue video clips, while \textbf{CMU-MOSEI} includes $22,856$ movie review clips sourced from YouTube. Both \textbf{DFEW} and \textbf{MER2023} consist of video clips from movies and TV series, each annotated with discrete emotion categories. In MSA, each sample is labelled with a sentiment score ranging from -$3$ to +$3$, representing seven sentiment levels. 
For MER, samples are labelled with six emotion classes.

\paragraph{Implementation Details.}
Our experiments were conducted using the PyTorch  \cite{pytorch} on a single $\operatorname{NVIDIA\space RTX} 4090$ GPU with 24GB memory. We set the initial learning rate, batch size, and decay interval to $1$e-$4$, $16$, $0.005$, respectively, and trained the model for $30$ epochs using the Adam optimizer.  For the initialization epoch and the hyperparameters related to consistency, we set: $\lambda = 5 $, $\rho = 4$, $k =0.5$, and $t=1/5$. We use Bert-base-uncased \cite{BERT} to obtain the original text features. We use Facet \cite{Facet} and COVAREP \cite{COVAREP} as visual encoder and acoustic encoder respectively. For the MSA task, we use accuracy of $2$ class (Acc$2$), accuracy of $7$ class (Acc$7$) and F$1$ scores as evaluation metrics, where Acc$2$ only focuses on the positive and negative for sentiment, while for the MER task, we use Unweighted Average Recall (UAR) and Weighted Average Recall (WAR) scores.  
\begin{table*}[tb]
    \centering
    \small
    \setlength{\tabcolsep}{11pt}
    \captionsetup{skip=8pt}
    
    \begin{tabular}{l|c|ccc|ccc}
            \toprule
            \multirow{2}{*}{\textbf{Model}}&\multirow{2}{*}{\textbf{Type}} & \multicolumn{3}{c|}{\textbf{MOSI}} & \multicolumn{3}{c}{\textbf{MOSEI}} \\ \cmidrule{3-8}
            ~         &~              & Acc-2 & F1  & Acc-7 & Acc-2 & F1 & Acc-7\\ 
            \midrule
            AGM$^\star$ (\textit{ICCV'23}) &   \multirow{4}*{ML}  & 78.23& 77.56&31.48& 72.14& 71.46&41.04\\
            ReconBoost$^\star$ (\textit{ICML'24})&~  & 77.96& 76.89&32.73& 68.61& 68.54&39.57\\
            LFM$^\star$ (\textit{NIPS'24})&~  &  79.41& 79.60&34.02&  74.88& 74.27&42.59\\
            CGGM$^\star$ (\textit{NIPS'24})&~  & 82.84& 82.74&33.73& 79.67& 78.97&45.76\\
            \midrule
            PMR (\textit{CVPR'21})&\multirow{5}*{MSA} & 83.60& 83.40&40.60& 83.30& 82.60&52.20\\ 
            FDMER (\textit{MM'22})&~    & 84.60& 84.70&44.10& 86.10& 85.80&54.10\\
            ConFEDE (\textit{ACL'23}) &~  & 85.52 & 85.52 &42.27& 85.82 & 85.83&54.86\\
            DMD (\textit{CVPR'23})&~ & 86.00& 86.00&45.60& 86.60& 86.60&54.50\\
            CAGC (\textit{CVPR'24})  &~      &  85.70&  85.60& 44.80& -& -&-\\
            \midrule
            TiCAL (Ours)&MSA   & \textbf{88.10}& \textbf{88.09}&\textbf{46.79}& \textbf{87.03} & \textbf{87.05} &\textbf{55.23}\\
            \bottomrule
            \end{tabular}
        \caption{Experimental results of state-of-the-art multimodal sentiment analysis (MSA) and multimodal learning (ML) models on two benchmark datasets: \textbf{MOSI} and \textbf{MOSEI}. Models marked with $^\star$ are reproduced using the publicly available code.}
    \label{tab:msaresult}
\end{table*}
\subsection{4.2 Comparison with State-of-the-Art Methods}\label{sec:mainresults}
We compare our model to three types of models: MSA or MER models, multimodal learning models, and language models (LLMs). 
The SOTA MSA models include CAGC \cite{CAGC}, DMD \cite{DMD}, PMR \cite{PMR}, FDMER \cite{FDMER}, and ConFEDE \cite{ConFEDE},  while the SOTA MER model includes Emotion-LLaMA \cite{Emotionllama}, VideoMAE \cite{VideoMAE}, MAE-DFER \cite{MAEDFER}, Former-DFER \cite{FormerDFER}, FBP \cite{FBP}, MER-Baseline \cite{Mer2023}. Specifically, for the MSA task, we compare with the most advanced multimodal learning methods, namely CGGM \cite{CGGM}, ReconBoost \cite{Reconboosticml2024}, AGM \cite{Boostingcvpr2023}, and LFM \cite{LFM}. In addition, we have also selected advanced and well-known LLMs for comparison:  GPT-4V \cite{GPT4v}, Video-Llama \cite{Videollama}, MiniGPT-v2 \cite{minigptv2}, LLaVA-NEXT \cite{llavanext}, and Qwen-Audio \cite{qwen}.

\paragraph{Multimodal Sentiment Analysis Results.}
Tab. \ref{tab:msaresult} presents the experimental results for the MSA task,  where the baseline method contains two types: multimodal learning (ML) and task-specific multimodal sentiment analysis (MSA) . The results indicate that TiCAL outperforms all previous models. For the MOSI dataset, our method achieves an average improvement of $2.1$\%, $2.09$\%, $1.19$\% over the SOTA MSA model DMD in terms of Acc-2, F1, and Acc-7, respectively. Our method on MOSEI also outperforms the baseline model, apart from F1. We attribute this improvement to the dynamic integration of multi-stage fusion in TiCAL, which leverages both the temporal and contextual dependencies more effectively. These results highlight the superiority of our method in multimodal sentiment analysis tasks.

\begin{table}[t]  
    \centering
    \small
        \centering
        \captionsetup{skip=8pt}

        \setlength{\tabcolsep}{10pt}
        \begin{tabular}{l|c|cc}
            \toprule
            \textbf{Method} &\textbf{Type}& \textbf{UAR} & \textbf{WAR} \\
            \midrule
            Qwen-Audio  &\multirow{5}*{LLM}               & 25.23    & 31.74 \\
            LLaVA-NEXT &~              & 25.12    & 33.75 \\
            MiniGPT-v2&~              & 35.36    & 34.47 \\
            Video-Llama &~             & 26.17    & 35.24 \\
            GPT-4V&~                 & 26.77    & 35.75 \\
            \midrule
            Former-DFER &\multirow{4}*{MER}             & 53.69    & 65.70 \\
            MAE-DFER  &~              & 60.99    & 74.43 \\
            VideoMAE  &~               & 63.41    & 74.60 \\
            Emotion-LLaMA &~         & 64.21    & 77.06 \\
            \midrule
            TiCAL (Ours)  &MER  & \textbf{65.87}    & \textbf{78.29} \\
            \bottomrule
        \end{tabular}
        \captionof{table}{Experimental results of MER on the LLM and MER models on DFEW dataset.}
        \label{tab:MERresult}
\end{table}
\begin{table}[t]
        \small
        \centering
        \captionsetup{skip=8pt}

        \setlength{\tabcolsep}{10pt}
        \begin{tabular}{l|c|c}
            \toprule
            \textbf{Method} & \textbf{Modality} & \textbf{F1} \\ \midrule
            MER2023-Baseline& A, V & 86.75 \\
            MER2023-Baseline & A, V, T & 86.40 \\
            FBP            & A, V, T & 88.55 \\
            VAT             & A, V, T & 89.11 \\
            Emotion-LLaMA        & A, V & 89.05 \\
            Emotion-LLaMA      & A, V, T & 90.36 \\ 
            \midrule
            TiCAL (Ours)         & A, V, T & \textbf{91.56} \\ 
            \bottomrule
        \end{tabular}
        \captionof{table}{ The experimental results of MER on the MER2023 dataset in terms of F1 score. Notably, the ``Modality'' column indicates the input modality configuration, where L, A, and V represent the language, audio, and visual modalities.}
        \label{tab:MERresult2023}
\end{table}

\paragraph{Multimodal Emotion Recognition Results.}
Tab.~\ref{tab:MERresult} and \ref{tab:MERresult2023} present the experimental results on the MER task, following the evaluation protocol of Emotion-LLaMA \cite{Emotionllama}. Our proposed method achieves new SOTA performance, surpassing both large language model (LLM)-based approaches and task-specific baselines across both the DFEW and MER2023 datasets. These empirical results demonstrate that TiCAL effectively captures and integrates multimodal emotional cues, whether for polarity-based sentiment classification or discrete emotion category recognition. This highlights the model's strong adaptability and its capacity to perceive nuanced emotional signals across modalities.


\subsection{4.3 Ablation Study}\label{sec:ablation}
\begin{table}[ht]
    \centering
    \small
    \setlength{\tabcolsep}{3pt}
    \begin{tabular}{l|ccc}
        \toprule
        \textbf{Components} &Acc-2 & F1  &Acc-7\\
        \midrule
        TiCAL  & \textbf{88.10}  &\textbf{88.09} &\textbf{46.79}\\
        \midrule
        Eq. \ref{eq:acloss} w/o $\boldsymbol{\tau}$ & 87.18(\textcolor[HTML]{199911}{-0.92}) & 87.20(\textcolor[HTML]{199911}{-0.89})  &44.77(\textcolor[HTML]{199911}{-1.02})\\
        Eq. \ref{eq:taskloss} w/o $\boldsymbol{\kappa}$ & 86.59(\textcolor[HTML]{199911}{-1.51}) & 86.55(\textcolor[HTML]{199911}{-1.54})   &44.68(\textcolor[HTML]{199911}{-2.11})\\  
        \midrule
        Eq. \ref{eq:typicality} w $d_{\mathbb{R}}$   & 87.24 (\textcolor[HTML]{199911}{-0.86}) & 87.18(\textcolor[HTML]{199911}{-0.91})
        &45.17(\textcolor[HTML]{199911}{-1.62})\\
        \midrule
        Eq. \ref{eq:allloss} w/o $\mathcal{L}_{Hyp}$   & 86.03(\textcolor[HTML]{199911}{-2.07}) & 86.17(\textcolor[HTML]{199911}{-1.92})   &44.24(\textcolor[HTML]{199911}{-2.55})\\
        \bottomrule
    \end{tabular}
    \captionof{table}{Ablation study results of the key components. ``w/o $\boldsymbol{\tau}$'' and ``w/o $\boldsymbol{\kappa}$'' represent removing typicality and consistency, respectively. $\mathcal{L}_{Hyp}$ is $HypCPCC$ loss in Sec. ~3.1. $d_{\mathbb{R}}$ is Euclidean distance.}
    \label{tab:abkey}
\end{table}
\paragraph{Effectiveness of Key Components.}
We perform ablation studies on the MOSI dataset to evaluate the contributions of individual components in our TiCAL framework. The results are presented in Tab.~\ref{tab:abkey}. First, removing the typicality-based dynamic optimization (see Sec.~3.3) results in a noticeable performance drop, indicating its importance in guiding modality integration. Excluding the consistency module leads to a further decrease in accuracy, demonstrating its critical role in capturing inter-modal alignment. Moreover, replacing the hyperbolic distance used in HASL with Euclidean distance degrades model performance, highlighting the effectiveness of hierarchical feature regularization in hyperbolic space. Finally, delete the unimodal optimization strategy (see Sec.~3.3) also negatively impacts performance, confirming the necessity of Estimation optimization for pseudo labels. More ablation experiments can be found in Appendix 3.1.
\begin{table*}[ht!]
    \centering
    \setlength{\tabcolsep}{3pt}
    \renewcommand{\arraystretch}{1.1}
    \captionsetup{skip=8pt}
    \begin{tabular}{c|c|c}
    \toprule
    \textbf{Input}
    &\begin{subtable}[b]{0.7\columnwidth}
		\centering
		\raisebox{-.5\height}{\includegraphics[width=\linewidth]{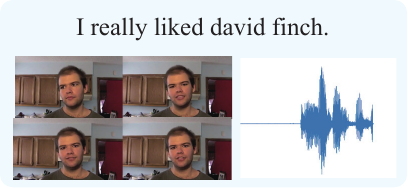}}\end{subtable} 
    &\begin{subtable}[b]{0.7\columnwidth}
		\centering
		\raisebox{-.5\height}{\includegraphics[width=\linewidth]{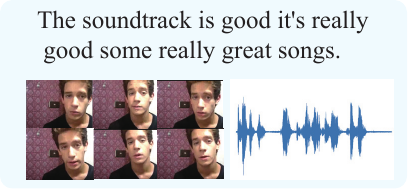}}\end{subtable} \\
    \midrule
    \textbf{DMD} \cite{DMD} & \textbf{Positive} \textbf{+3} \textcolor{red}{\ding{55}} & \textbf{Negative} \textbf{-1} \textcolor{red}{\ding{55}} \\

    \textbf{CAGC}  \cite{CAGC}& \textbf{Positive} \textbf{+1} \textcolor{green}{\ding{51}}& \textbf{Positive} \textbf{+1}  \textcolor{red}{\ding{55}} \\
    \midrule
    \multirow{4}{*}{\textbf{TiCAL}}&$\boldsymbol{\tau_l}=0.91,\boldsymbol{\tau_v}=0.73,\boldsymbol{\tau_a}=0.61$&$\boldsymbol{\tau_l}=0.68,\boldsymbol{\tau_v}=0.21,\boldsymbol{\tau_a}=0.11$\\    ~&$\boldsymbol{y_l^*}=+2,\boldsymbol{y_v^*}=+1,\boldsymbol{y_a^*}=+1$&$\boldsymbol{y_l^*}=+2,\boldsymbol{y_v^*}=-1,\boldsymbol{y_a^*}=-1$\\
    ~&$\boldsymbol{\kappa}=0.8173$&$\boldsymbol{\kappa}=0.4712$\\
    ~& \textbf{Positive} \textbf{+1}\textcolor{green}{\ding{51}} & \textbf{Positive} \textbf{+2} \textcolor{green}{\ding{51}} \\
    \bottomrule
    \end{tabular}
    \caption{Two real-case examples from the MOSI test set comparing predictions across different state-of-the-art MSA models and our proposed TiCAL method. Each example includes visualized input modalities to aid interpretability parameters.  These examples highlight TiCAL’s ability to make accurate predictions, even when baseline models fail, particularly in the presence of modality conflicts or ambiguous cues.}
    \label{fig:casestudy}
\end{table*}

 \begin{figure}[t]
    \centering
    \begin{subfigure}{0.22\textwidth}
      \centering
      \includegraphics[width=\textwidth]{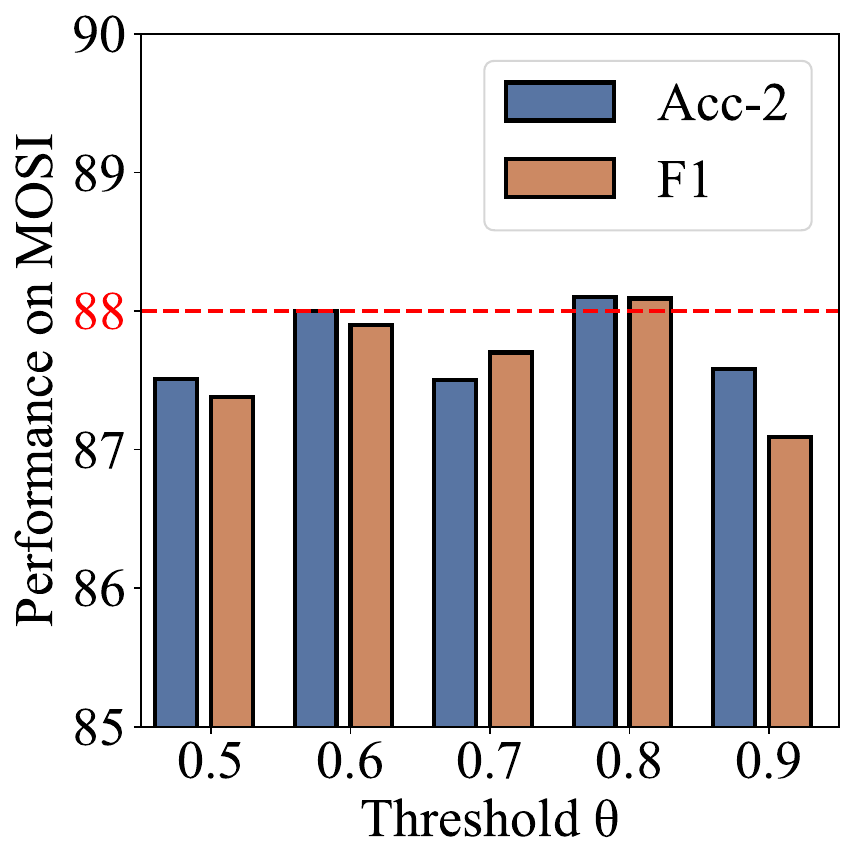}
    \end{subfigure}
    \hfill
    \begin{subfigure}{0.22\textwidth}
      \centering
      \includegraphics[width=\textwidth]{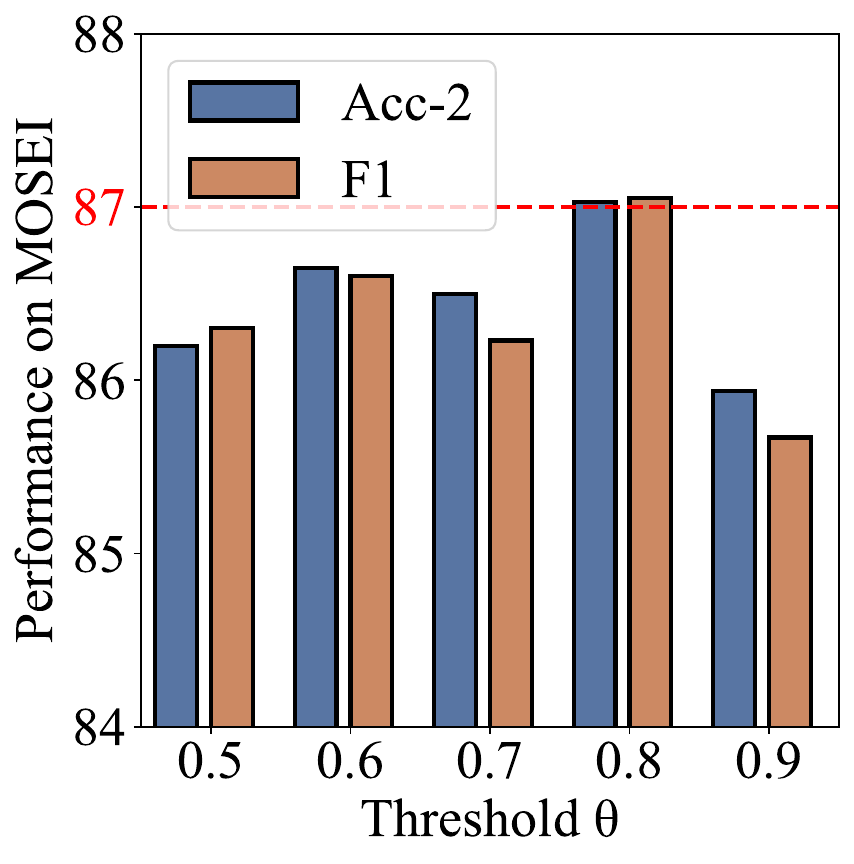}
    \end{subfigure}
    \caption{The performance of our TiCAL method with different confidence threshold $\theta$ in HASL initialization. The left side is on MOSI and the right side is on MOSEI.}
     \label{fig:abthreshold}
\end{figure}

\paragraph{Impact of Confidence Threshold.}
To examine the effect of the threshold  $\theta$ on model performance, we evaluate TiCAL under five different $\theta$ values and report the results on the MOSI and MOSEI dataset in Fig.~\ref{fig:abthreshold}. The results show that the model achieves the highest accuracy when $\theta = 0.8$. Intuitively, a larger $\theta$ enforces higher typicality among stored samples. However, when $\theta = 0.9$, the threshold may become overly restrictive,  preventing proper initialization of the HASL module, which in turn leads to performance degradation. More hyperparameter analyses are provided in Appendix 3.2.

\subsection{4.4 Case Analysis}
As illustrated in Table~\ref{fig:casestudy}, we present a case study by visualizing representative examples of both successful and failed predictions made by different multimodal sentiment analysis (MSA) methods on the MOSI dataset. For each sample, we include the pseudo unimodal labels, their corresponding typicality scores, and the consistency measure computed using our proposed framework.
In the first example, where the multimodal signals are relatively consistent, DMD fails to make a correct prediction, while DMD succeeds. However, in the second example—characterized by significant cross-modal inconsistency—both DMD and CAGC fail to classify the emotion correctly. In contrast, our TiCAL framework successfully predicts both samples, regardless of their consistency level.

These examples highlight TiCAL's ability to adaptively interpret unimodal cues and leverage typicality-weighted consistency to guide learning. This case study underscores the model’s robustness in handling emotionally conflicting modalities, which is critical for real-world applications of multimodal emotion recognition.

\section{5 Conclusion}

In this work, we proposed TiCAL (Typicality-based Consistency-Aware Learning), a novel framework inspired by human emotion perception mechanisms, to address the long-overlooked issue of inter-modal emotional conflicts in multimodal emotion recognition (MER). By generating pseudo unimodal emotion labels and embedding them in a hyperbolic space, TiCAL captures modality-specific emotional tendencies with structured semantic representation. Moreover, our typicality-based consistency metric enables dynamic, stage-wise optimization that reflects human-like emotional integration. Extensive experiments across diverse MER and MSA benchmarks confirm the effectiveness of TiCAL, achieving state-of-the-art performance while offering greater interpretability and robustness in handling conflicting modality cues.

Despite these promising results, our work has several \textbf{limitations}. For example, TiCAL assumes full modality availability during both training and inference, which may not be practical in real-world applications where certain modalities can be noisy or missing. 
Future work will explore the extension of TiCAL to handle incomplete modalities, enabling developing more general and robust multimodal emotion recognition models. 

\section*{Acknowledgments}
This research was partially supported by the National Natural Science Foundation of China (NSFC) (62306064 and U19A2059) and the Sichuan Science and Technology Program (2022ZHCG0008, 2023ZYD0165 and 2024ZDZX0011).   We appreciate all the authors for their fruitful discussions.
In addition, thanks are extended to anonymous reviewers for their insightful comments.

\bigskip
\bibliography{aaai2026}
\end{document}